%% file: main_round_2.tex
\documentclass[10pt,twocolumn,letterpaper]{article}

\usepackage[accsupp]{axessibility} 
\usepackage[pagenumbers]{wacv} 

\input{preamble}

\definecolor{wacvblue}{rgb}{0.21,0.49,0.74}
\usepackage[pagebackref,breaklinks,colorlinks,allcolors=wacvblue]{hyperref}

\def\wacvPaperID{1594} 
\def\confName{WACV}
\def\confYear{2027}

\title{ReVA-HPE: Relative 6DoF Head Pose Estimation with Synthetic Pose Supervision and Personalized Virtual Anchors}

\author{Vasiliki Vasileiou$^{1,2}$ \quad
        Panagiotis P. Filntisis$^{1}$ \quad
        Petros Maragos$^{1,2}$ \quad
        Kostas Daniilidis$^{3,4}$ \vspace{2pt}\\
  \small{$^1$HERON - Hellenic Robotics Center of Excellence, Athena Research Center, Greece}\vspace{-2pt}\\
  \small{$^2$ National Technical University of Athens, School of Electrical and Computer Engineering, Greece}\vspace{-2pt}\\
  \small{$^3$Robotics Institute, Athena Research Center, Greece} \quad
  \small{$^4$University of Pennsylvania, USA}
  \vspace{-2pt}\\
  {\tt\footnotesize \{v.vasileiou, pfilntisis, petros.maragos, k.daniilidis\}@athenarc.gr} \vspace{-6pt}
}

\begin{document}
\maketitle
\input{sec_round_2/0_abstract}
\input{sec_round_2/1_intro}
\input{sec_round_2/2_related_Vas}
\input{sec_round_2/3_method_VAs}
\input{sec_round_2/4_experiments}
\input{sec_round_2/7_conclusion}
{
    \small
    \bibliographystyle{ieeenat_fullname}
    \bibliography{main}
}

\end{document}

%% file: preamble.tex
\usepackage{amsmath,amssymb,amsfonts,mathtools}
\usepackage{booktabs,multirow,tabularx,array}
\usepackage{graphicx}
\usepackage[table]{xcolor}
\usepackage{microtype}
\usepackage{enumitem}
\usepackage{algorithm}
\usepackage[noend]{algpseudocode}
\usepackage{tikz}
\usetikzlibrary{arrows.meta,positioning,fit,calc,shapes.geometric,backgrounds,decorations.pathreplacing}
\usepackage{pgfplots}
\pgfplotsset{compat=1.18}
\usepackage{siunitx}
\usepackage{adjustbox}
\usepackage{makecell}
\usepackage{subcaption}
\usepackage{url}
\usepackage{bm}
\usepackage{xspace}
\usepackage{wrapfig}
\newsavebox{\acklogo}

\definecolor{anchorblue}{RGB}{50,105,168}
\definecolor{avatarorange}{RGB}{238,140,48}
\definecolor{metricgreen}{RGB}{72,154,104}
\definecolor{querypurple}{RGB}{132,91,166}
\definecolor{syntheticteal}{RGB}{38,145,153}
\definecolor{lightgray}{RGB}{245,246,248}
\definecolor{darkgray}{RGB}{70,74,82}
\definecolor{todored}{RGB}{185,35,35}

\newcommand{\vggthpe}{\textsc{VGGT-HPE}\xspace}

\newcommand{\soThree}{\mathrm{SO}(3)}
\newcommand{\seThree}{\mathrm{SE}(3)}
\newcommand{\transl}{\mathbf{t}}
\newcommand{\rot}{\mathbf{R}}
\newcommand{\pose}{\mathbf{P}}

\newcommand{\famos}{\textsc{FaMoS}\xspace}

\setlist[itemize]{leftmargin=*,topsep=2pt,itemsep=1pt,parsep=0pt}
\setlist[enumerate]{leftmargin=*,topsep=2pt,itemsep=1pt,parsep=0pt}

%% file: sec_round_2/0_abstract.tex
\begin{abstract}
Monocular head pose estimation is usually formulated as absolute regression
from a single face crop, yet many applications observe the same person over
time and can supply an explicit anchor image. \textcolor{black}{Recently, relative head pose estimation was proposed, which casts the problem as estimation of the relative geometric change between an anchor and a query image}. However, existing methods still require the anchor image pose to be known at inference, significantly limiting their practical use. We first
extend relative head pose estimation to full 6DoF by additionally predicting a depth-normalized relative translation, which composes
with the anchor to recover metric query pose in the camera frame.
To eliminate the need for a ground-truth anchor pose, we reconstruct a personalized avatar from a single unlabeled enrollment image and render a bank of virtual anchors with known poses. The bank provides robust rotation hypotheses through multi-anchor consensus, while the real enrollment image serves as the real anchor in the translation branch via landmark PnP. We also introduce an efficient variant that amortizes the bank cost once at enrollment, after which each query requires a single real-to-real relative prediction, with limited accuracy sacrifice. 
\textcolor{black}{The relative pose estimator is trained exclusively on synthetic face renderings, without real pose supervision, pseudo-labels, domain adaptation, or fine-tuning on the evaluation sets. On real 6DoF benchmarks, ReVA-HPE achieves state-of-the-art rotation accuracy on BIWI and, given a near-frontal enrollment image, on FaMoS, while remaining competitive in metric translation despite relying solely on synthetic pose supervision. Project page and code: \url{https://vasilikivas.github.io/ReVA-HPE/}}
\end{abstract}
\vspace{-0.28cm}

%% file: sec_round_2/1_intro.tex
\section{Introduction}
\label{sec:intro}
Head pose provides a compact geometric cue to a person's attention and interaction with the surroundings. Reliable estimation of this cue supports applications including gaze analysis, driver monitoring, facial understanding, telepresence, sign-language systems, and human-robot interaction~\citep{murphy2009headpose,abate2022headpose}.
\begin{figure}[t]
  \centering
  \includegraphics[width=0.85 \columnwidth]{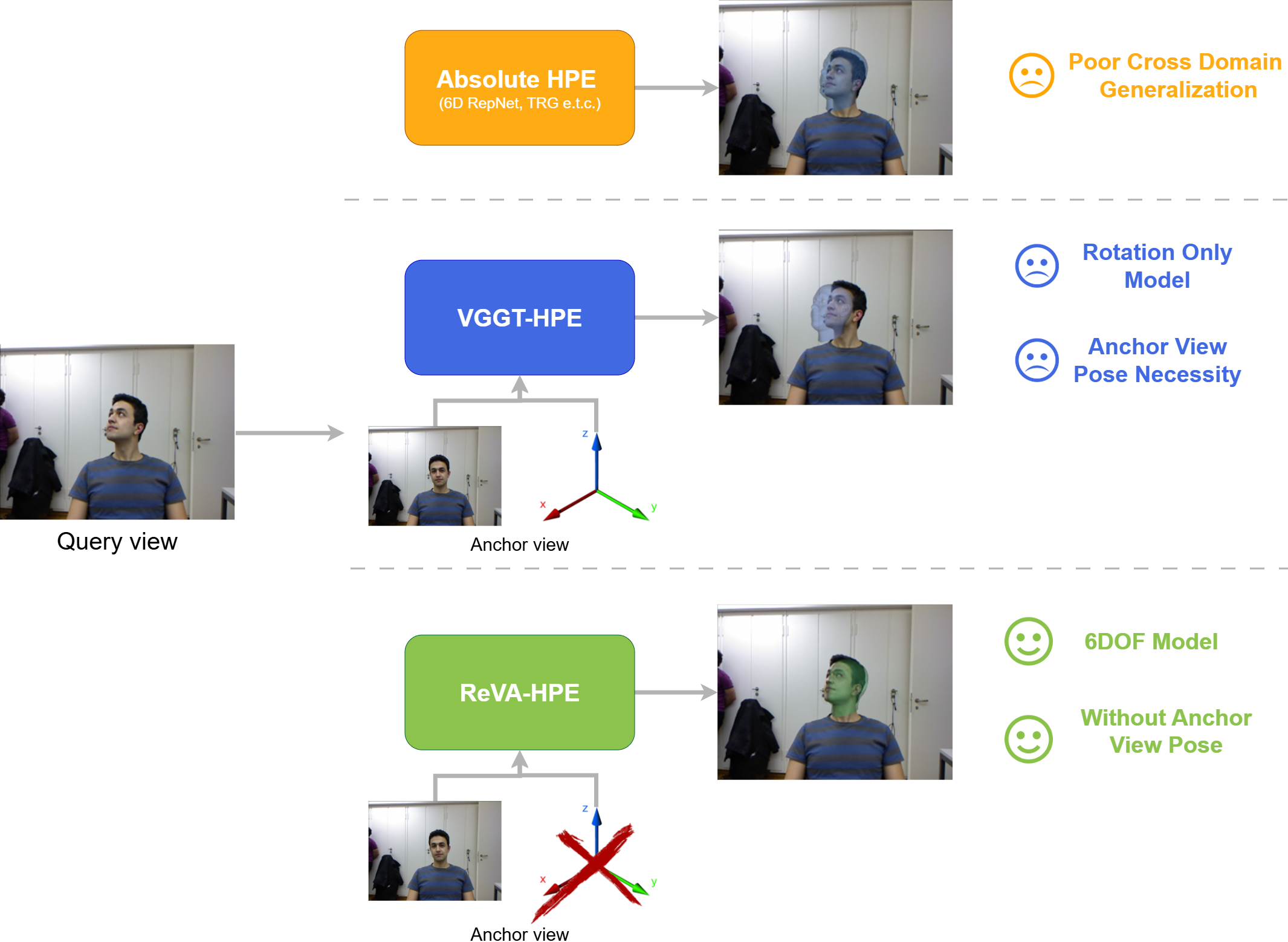}
  \vspace{-0.2cm}
  \caption{
  Absolute HPE predicts pose directly from the query and may generalize poorly across domains. VGGT-HPE uses relative prediction but requires a reference image with known pose and estimates rotation only. ReVA-HPE removes this requirement, using personalized virtual anchors for rotation and PnP on an unlabeled enrollment image for translation, and estimates full 6DoF head pose without any ground-truth pose.}
  \label{fig:teaser}
  \vspace{-0.41cm}
\end{figure}

\vspace{-12pt}
Most learned head pose estimators focus on orientation, predicting yaw, pitch, and roll- or an equivalent rotation representation-directly from a single face crop~\citep{ruiz2018fine,zhou2020whenet,hempel2022sixdrepnet,zhang2023tokenhpe}. A smaller body of work targets full 6DoF pose, additionally recovering the metric translation of the head with respect to the camera~\citep{albiero2021img2pose,kao2023perspnet,chun2024trg}. This setting is considerably more challenging: rotation must remain robust to extreme poses and appearance variation, while translation also depends on camera intrinsics, crop geometry, subject scale, and the precise definition of the anatomical head origin.

\textcolor{black}{Recently, \vggthpe~\citep{vasileiou2026vggt} reformulated head pose estimation as relative pose prediction: given an anchor and a query image, the model estimates their relative geometric transformation and composes it with the anchor pose to recover the query pose. This image-matching formulation enabled \vggthpe to outperform leading absolute-regression methods on real-face benchmarks, including methods using real or mixed pose supervision, despite relying solely on synthetic pose supervision.}

However, two limitations remained. First, \vggthpe was evaluated mainly as an
orientation estimator. A practical 6DoF system also needs a well-defined
relative translation target, correct metric decoding, and camera-frame
translation evaluation. Second, its strongest protocol still assumed a known
anchor pose. Even one ground-truth anchor pose per sequence is an oracle, yet the
obvious fix is self-defeating: an absolute estimator accurate enough to supply
that would already solve the problem that motivates the relative
formulation. Replacing the ground truth anchor with a weaker absolute estimate
injects a persistent offset into every composed query, while autoregressive
anchoring drifts over time.

In this paper, we turn relative head pose estimation into a complete 6DoF system that does not require a ground-truth anchor pose, called Relative Virtual-Anchor Head Pose Estimation (ReVA-HPE) and summarized in Figure \ref{fig:teaser}. We first extend the previous rotation-centered formulation to predict both relative rotation and relative translation, allowing the query head pose to be recovered metrically in the camera frame. We then remove the known-anchor-pose assumption by constructing, from a single unlabeled enrollment image, a personalized virtual-anchor bank with known poses. We estimate the query rotation by robustly combining predictions from the virtual anchors in the bank. Metric translation is obtained separately through a single PnP solve on the enrollment image.
\textcolor{black}{We also introduce an efficient variant that processes the virtual-anchor bank only once, during enrollment. After this one-time calibration, each new query requires a single network inference, with only a small loss in accuracy. The relative pose estimator is trained exclusively on procedural face renderings, without real pose supervision, pseudo-labels, or target-domain fine-tuning. Despite this restricted supervision, ReVA-HPE achieves state-of-the-art rotation accuracy on BIWI and, given a near-frontal enrollment image, on FaMoS, while remaining competitive in metric translation.
}Our contributions are:


\begin{itemize}
\item \textbf{Relative 6DoF head pose estimation.} We extend the previously proposed relative head pose rotation formulation to full 6DoF by learning relative translation together with relative rotation and decoding metric query pose in the camera frame.
\item \textbf{Ground-truth-free virtual anchors.} We remove the ground-truth anchor pose requirement by replacing the labeled real anchor with a personalized virtual anchor set constructed from one unlabeled enrollment image.
\item \textbf{Efficient enrollment calibration.} We introduce a lightweight variant that uses the virtual anchor bank once at enrollment to calibrate its rotation, combines it with a PnP-grounded translation estimate, and processes every subsequent query with a single real-to-real relative prediction.
\end{itemize}

\begin{figure*}[t]
  \centering
  \includegraphics[width=0.91\textwidth]{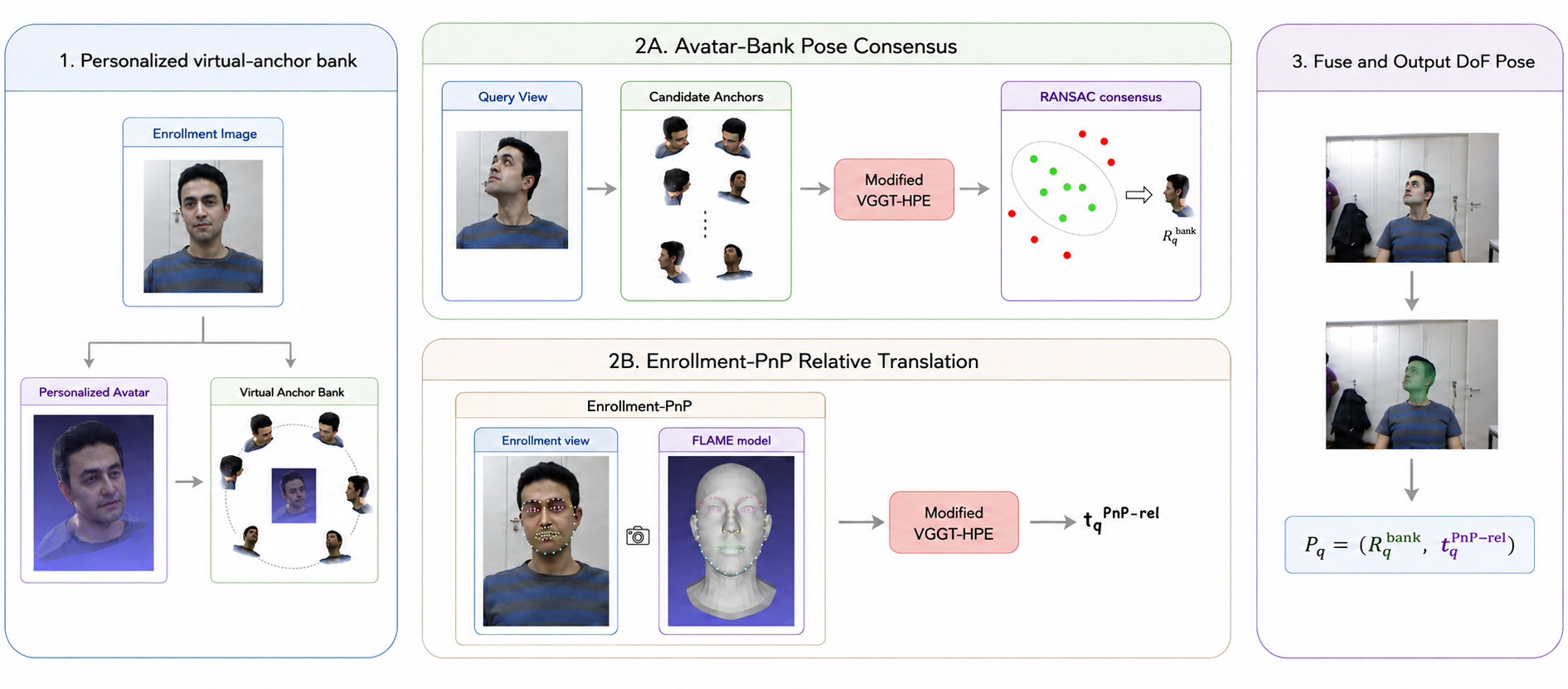}
    \vspace{-0.3cm}
  \caption{\textbf{ReVA-HPE overview.} From a single unlabeled enrollment
  image, a personalized Gaussian avatar is reconstructed and rendered at $M$
  known poses to form the virtual anchor bank. \textbf{(2A)} For each query,
  the relative 6DoF model compares the query against all bank anchors,
  producing $M$ rotation candidates that are fused by pose-space RANSAC
  consensus into $\widehat{\rot}^{\mathrm{bank}}_q$. \textbf{(2B)} In
  parallel, 68 landmarks detected on the enrollment image ground a metric
  PnP solve. The relative model then predicts the enrollment-to-query
  translation update, yielding
  $\widehat{\transl}^{\,\mathrm{PnP\text{-}rel}}_q$. 
  Modified VGGT-HPE denotes the frozen VGGT camera branch adapted with LoRA and trained to predict relative rotation together with the depth-normalized translation update \((\delta_x,\delta_y,\delta_z)\) from an ordered image pair. The two components
  are fused into the final 6DoF pose
  $\widehat{\pose}_q = (\widehat{\rot}^{\mathrm{bank}}_q,\,
  \widehat{\transl}^{\,\mathrm{PnP\text{-}rel}}_q)$
  without any ground-truth anchor pose at inference.}
  \label{fig:pipeline}
  \vspace{-0.3cm}
\end{figure*}

%% file: sec_round_2/2_related_Vas.tex
\section{Related Work}
\label{sec:related}

\subsection{Monocular Head Pose and 6DoF Face Pose}

Early head-pose systems relied on facial correspondences, 3D templates, and perspective pose solvers. Modern approaches predominantly learn an absolute mapping from a face crop to Euler angles or a rotation representation. HopeNet combines classification and regression~\citep{ruiz2018fine}, \mbox{FSA-Net} aggregates fine-grained facial structure~\citep{yang2019fsanet}, WHENet targets wide-range orientation~\citep{zhou2020whenet}, 6DRepNet uses a continuous 6D rotation representation with geodesic supervision~\citep{hempel2022sixdrepnet}, and TokenHPE learns transformer orientation tokens~\citep{zhang2023tokenhpe}. These methods are effective for orientation but learn a dataset-specific absolute reference from training images.

Fewer methods evaluate complete 6DoF head pose. img2pose directly predicts face pose and detection~\citep{albiero2021img2pose}, \mbox{PerspNet} reconstructs facial geometry under perspective projection and recovers 6DoF pose~\citep{kao2023perspnet}, and TRG explicitly couples head translation, rotation, and facial geometry~\citep{chun2024trg}. Landmark PnP remains a strong geometric baseline when calibrated intrinsics and reliable 2D-3D correspondences are available~\citep{lepetit2009epnp,fischler1981ransac}. Our method differs in predicting both rotation and translation \emph{relative to a visible anchor}, then using virtual anchors to remove the need for a GT pose.

Recent methods address robustness and uncertainty in complementary ways.
DSFNet~\citep{li2023dsfnet} fuses image and 3D representation spaces for
occlusion-robust dense face alignment, while
HeadDiff~\citep{wang2024headdiff} models uncertainty in head rotation using a
diffusion process. OPAL~\citep{cobo2024opal} studies rotation
representations across different pose ranges, and
SemiUHPE~\citep{zhou2026semiuhpe} uses semi-supervised learning to improve
unconstrained head-pose estimation. These methods primarily target rotation
or facial alignment, whereas ReVA-HPE estimates metric 6DoF pose relative
to an explicit visual anchor and removes the need for a labeled anchor.

\subsection{Relative Pose \& Geometry Foundation Models}

Relative transformations are standard in visual odometry, SLAM, tracking, and reference-based object pose estimation because correspondences across observations constrain displacement directly~\citep{teed2021droid,wang2021tartanvo}. Recent geometry foundation models generalize this idea across broad image domains. CroCo learns cross-view completion~\citep{weinzaepfel2022croco}, DUSt3R predicts point maps without known calibration~\citep{wang2024dust3r}, MASt3R grounds feature matching in 3D~\citep{leroy2024mast3r}, and VGGT jointly predicts cameras and dense scene geometry from one or more images~\citep{wang2025vggt}. \vggthpe~\citep{vasileiou2026vggt} adapts VGGT's camera branch to paired head images with LoRA. In this work we keep the relative image-pair formulation, extend it to explicit relative translation, and address the test-time anchor problem by constructing RGB-derived anchor sources from a single enrollment image.



\subsection{\textcolor{black}{Synthetic Pose Supervision}}

Synthetic supervision in head pose estimation commonly relies on
pose-expanded real images. For example, 300W-LP~\citep{zhu2016facealignment}
uses 3D face profiling to generate large-pose views from real photographs
and is widely used to train absolute estimators such as
6DRepNet~\citep{hempel2022sixdrepnet} and
TokenHPE~\citep{zhang2023tokenhpe}. Procedural face models provide a
different form of supervision: FLAME~\citep{li2017flame} allows direct
control over identity, expression, viewpoint, and pose without deriving
training examples from real images. VGGT-HPE~\citep{vasileiou2026vggt}
showed that relative rotation learned from such procedural renderings
transfers effectively to real faces. We build on this setting, extending
the relative target to full 6DoF while removing the need for a known anchor
pose. The task-specific adaptation of our relative estimator uses no real
head-pose images, pose annotations, pseudo-labels, or target-domain
fine-tuning.



\subsection{Personalized Avatars as Virtual References}

Recent 3D Gaussian-splatting methods have enabled controllable 3D
human-head avatars and high-quality novel-view synthesis~\citep{kerbl2023gaussians}.
Single-image or few-image avatar models reconstruct personalized human heads
for view generation or animation, including FaceLift~\citep{lyu2024facelift},
LAM~\citep{he2025lam}, and SEGA~\citep{guo2025sega}. These works primarily
target visual fidelity, editability, and controllable rendering.

\textcolor{black}{In contrast, the present work uses a personalized avatar as a calibration
device for relative pose estimation. The reconstructed avatar renders
identity-consistent references with known synthetic poses, which provide
multiple anchor-conditioned rotation hypotheses. Metric translation is not
taken from the avatar; it is grounded by the real enrollment image through
PnP.}
\textcolor{black}{Importantly, FaceLift is trained entirely on synthetic data and is used off-the-shelf, without adaptation to the evaluation sets.}


%% file: sec_round_2/3_method_VAs.tex
\section{Method}
\label{sec:method}

We build on \vggthpe~\citep{vasileiou2026vggt}, which estimates head pose by predicting the geometric change between an anchor and a query image. Two components are missing for a practical 6DoF system. First, the original relative prediction covers only rotation. We add an explicit depth-normalized translation target, allowing us to recover metric camera-frame translation once the translation of the anchor is known. Second, \vggthpe requires the ground-truth pose of the anchor image at test time. We remove this requirement by replacing the labeled anchor with a single unlabeled enrollment image. From this image, we construct a personalized bank of virtual anchors for rotation and use landmark-based PnP to obtain the metric translation of the enrollment anchor. The full pipeline is illustrated in Figure~\ref{fig:pipeline}.

\subsection{VGGT-HPE Preliminaries}
\label{sec:prelim}

\vggthpe~\citep{vasileiou2026vggt} reframes head pose estimation as relative prediction. Instead of regressing an absolute pose from a single crop, it receives an anchor image and a query image and predicts the geometric change between them, which is then composed with the anchor image ground-truth pose. This replaces the implicit canonical frame learned by absolute regressors with an explicit visual reference. We keep the pairwise VGGT-HPE architecture and relative-rotation composition, and extend its camera target with explicit depth-normalized translation.

\noindent \textbf{Pose convention.}
We work in the RGB camera frame throughout. A head pose is
\begin{equation}
    \pose=(\rot,\transl),\qquad
    \rot\in\soThree,\quad
    \transl=[x,y,z]^\top\in\mathbb{R}^3,
\end{equation}
where \(\rot\) maps canonical head coordinates into the camera frame and
\(\transl\) locates the canonical head origin. The model always receives an
ordered pair: an anchor \(I_a\) with known pose \(\pose_a=(\rot_a,\transl_a)\)
and a query \(I_q\), and predicts the geometric change from anchor to query.

\noindent \textbf{Backbone and adaptation.}
We reuse VGGT's multi-view camera branch~\citep{wang2025vggt}, freezing the backbone and disabling the depth, point-map, and tracking heads.
Rank-8 LoRA adapters with scale 16~\citep{hu2022lora} are inserted into the
attention and MLP projections of the frame and global transformer blocks,
adapting a small number of parameters to facial geometry while preserving the
pretrained geometric prior.

\noindent \textbf{Relative rotation.}
From the adapted pair, the model predicts the anchor-to-query rotation
\begin{equation}
    \Delta\rot_{q\leftarrow a}=\rot_q\rot_a^\top,
    \label{eq:relrot}
\end{equation}
and recovers the query orientation by composing it with the anchor pose,
\begin{equation}
    \widehat{\rot}_q=
    \widehat{\Delta\rot}_{q\leftarrow a}\rot_a.
    \label{eq:rotdecode}
\end{equation}

\subsection{Relative 6DoF Pose Learning}
\label{sec:rel6dof}

\subsubsection{Depth-Normalized Relative Translation}
\label{sec:depthnorm}

Translation is less direct than rotation. A homogeneous relative-pose translation couples translation to the anchor rotation; with a distant head, small rotation errors can induce large apparent translation errors. Directly regressing \(\transl_q-\transl_a\) is also strongly scale-dependent. We therefore predict a depth-normalized displacement,
\begin{equation}
  \delta_x=\frac{x_q-x_a}{z_a},\qquad
  \delta_y=\frac{y_q-y_a}{z_a},\qquad
  \delta_z=\log\frac{z_q}{z_a}.
  \label{eq:deltatarget}
\end{equation}
The lateral components are normalized by anchor depth, while the log-depth ratio makes forward and backward motion approximately symmetric.

At inference, once an anchor translation
\(\transl_a=[x_a,y_a,z_a]^\top\) is available, the metric query translation is
decoded as
\begin{equation}
  \widehat z_q=z_a\exp(\widehat\delta_z),
  \label{eq:zdecode}
\end{equation}
\begin{equation}
  \widehat x_q=x_a+z_a\widehat\delta_x,\qquad
  \widehat y_q=y_a+z_a\widehat\delta_y.
  \label{eq:xydecode}
\end{equation}
The output
\((\widehat{\Delta\rot}_{q\leftarrow a},
\widehat{\delta}_x,\widehat{\delta}_y,\widehat{\delta}_z)\)
forms our relative 6DoF representation. Its translation component is not the translation block of a homogeneous \(\seThree\) product; Eqs.~\eqref{eq:zdecode}-\eqref{eq:xydecode} define the verified decoder used by all reported experiments.

\noindent \textbf{Branch-wise composition.}
Under Eqs.~\eqref{eq:rotdecode}-\eqref{eq:xydecode}, the anchor rotation enters only the rotation composition, while the anchor translation enters only the translation decoder. This separation is important in the ground-truth-free setting: the two anchor components may be estimated from different measurements as long as they are expressed in the same RGB camera coordinate frame. We use this property in Sec.~\ref{sec:dropanchor}: personalized virtual anchors provide robust rotation hypotheses, while the real enrollment image provides a metric translation gauge.

\subsubsection{Training Objective}
\label{sec:trainingobjective}

VGGT emits camera predictions at \(K\) intermediate stages. We supervise every stage with exponentially decayed weights,
\begin{equation}
\mathcal{L}_{\mathrm{cam}}=
\frac{1}{K}\sum_{k=1}^{K}\gamma^{K-k}
\left(
\lambda_R\mathcal{L}^{(k)}_R+
\lambda_T\mathcal{L}^{(k)}_T+
\lambda_F\mathcal{L}^{(k)}_F
\right),
\label{eq:loss}
\end{equation}
where \(\mathcal{L}_R\) is an \(\ell_1\) quaternion loss,
\(\mathcal{L}_T=\|\widehat{\boldsymbol\delta}-\boldsymbol\delta\|_1\), and \(\mathcal{L}_F\) supervises the inter-frame focal-length ratio in log-tangent space. We use \(\lambda_R=1\), \(\lambda_T=1\), and \(\lambda_F=0.5\). Joint camera supervision is retained because removing translation and field-of-view supervision degrades cross-domain rotational accuracy in the preliminary ablation~\citep{vasileiou2026vggt}.

\subsubsection{Controlled Synthetic Training Data}
\label{sec:synthetic}

The head-pose network is trained \emph{only} on procedural FLAME
renders~\citep{li2017flame}: 250 identities - 200 with synthetic hairstyles,
50 bald - rendered under two HDR environment maps, ten expressions, and five
viewpoints per expression, giving 25,000 base images across 54 skin-tone
textures. Pair members share identity and illumination but may differ in
expression and viewpoint. 

Compared with \vggthpe, we use a more controlled geometric protocol:
a fixed list of $10^6$ predefined anchor-query pairs instead of online
random sampling, fixed-focal copies of the data, and a unified pair crop
with margin $0.5$ and analytically updated intrinsics. Appearance
augmentation is retained; geometry-changing randomness from focal variation
and independent crop perturbations is removed. Only LoRA parameters are
optimized with AdamW and mixed precision, the VGGT backbone remains frozen. Examples of our synthetic data can be found in the Suppl. Mat.

\subsection{Removing the Ground-Truth Anchor Pose}
\label{sec:dropanchor}


The relative predictor requires an anchor pose to recover an absolute
query pose. The GT-anchor oracle provides the ground-truth pose of the
enrollment anchor from benchmark annotations, but such a protocol is
not deployable. We instead assume a single \textit{unlabeled} enrollment image
\(I_e\) for each subject. A personalized virtual-anchor bank provides
query-rotation hypotheses from anchors with known synthetic poses,
while enrollment PnP estimates the metric translation of the real
enrollment anchor. Our final method therefore obtains query rotation
through virtual-anchor consensus and query translation through
enrollment-to-query relative decoding.

\subsubsection{Avatar-Bank Pose Consensus}
\label{sec:bankrotation}
Anchor poses are known by construction, as each anchor is rendered from a viewpoint sampled over fixed front-hemisphere directions in canonical head coordinates. The avatar is registered to this frame at enrollment by triangulating 68 landmarks across calibrated rendered views and aligning them to the enrollment FLAME fit with a RANSAC-robust similarity transform. The Suppl. Mat. quantifies the alignment residual. For each query \(I_q\), we compare the query with every rendered virtual anchor:
\begin{equation}
  (\widehat{\Delta\rot}^{(m)}_{q\leftarrow m},
   \widehat{\boldsymbol\delta}^{(m)}_{q\leftarrow m})
  =
  f_\theta(A_m,I_q).
\end{equation}
Each virtual anchor gives one absolute 6DoF candidate,
\begin{equation}
  \widehat{\rot}^{(m)}_q
  =
  \widehat{\Delta\rot}^{(m)}_{q\leftarrow m}\rot_m,
  \qquad
  \widehat{\transl}^{(m)}_q
  =
  \operatorname{Decode}
  \left(
  \widehat{\boldsymbol\delta}^{(m)}_{q\leftarrow m},
  \transl_m
  \right).
  \label{eq:bank_pose_candidates}
\end{equation}

The bank candidates are fused by pose-consensus RANSAC. Each candidate \(i\)
is used as a seed. For every other candidate \(j\), we compute a rotation
residual and a translation residual,
\begin{equation}
  r_R^{(i,j)}
  =
  d_R
  \left(
  \widehat{\rot}^{(j)}_q,
  \widehat{\rot}^{(i)}_q
  \right),
  \qquad
  r_t^{(i,j)}
  =
  \left\|
  \widehat{\transl}^{(j)}_q
  -
  \widehat{\transl}^{(i)}_q
  \right\|_2 ,
\end{equation}
where \(d_R\) is geodesic rotation distance. Candidate \(j\) is an inlier to
seed \(i\) when
\begin{equation}
  r_R^{(i,j)} \leq \tau_R,
  \qquad
  r_t^{(i,j)} \leq \tau_t .
  \label{eq:bank_ransac_inlier}
\end{equation}

We use $\tau_R=10^\circ$, $\tau_t=50 \ mm$, and require at least three inliers. These values are fixed globally and are not retuned on BIWI or FaMoS. A synthetic-only sensitivity analysis, including corrupted anchor hypotheses, shows that this operating point lies within a stable region spanning approximately $\tau_R=7.5^\circ-15^\circ$ and $\tau_t\geq50 \ mm$, full results are provided in the supplementary material. In the final ReVA-HPE method, $\tau_t$ is used only to determine the internally consistent bank hypotheses contributing to rotation. Each seed i is scored first by its inlier count ${|\mathcal{I}_i|}$. Ties are broken by the mean normalized residual
\begin{equation}
  s_i = \frac{1}{|\mathcal{I}_i|}
  \sum_{j\in\mathcal{I}_i}
  \left(
  \frac{r_R^{(i,j)}}{\tau_R}
  +
  \frac{r_t^{(i,j)}}{\tau_t}
  \right).
\end{equation}



Let \(\mathcal{I}^{\star}\) be the best consensus set. The bank rotation is
the projected mean of the inlier rotations,
\begin{equation}
  \widehat{\rot}^{\mathrm{bank}}_q
  =
  \Pi_{\soThree}
  \left(
  \frac{1}{|\mathcal{I}^{\star}|}
  \sum_{j\in\mathcal{I}^{\star}}
  \widehat{\rot}^{(j)}_q
  \right),
  \label{eq:bank_projected_mean}
\end{equation}
where \(\Pi_{\soThree}\) denotes SVD projection onto \(\soThree\) \cite{doi:10.1137/S0895479801383877}.

\subsubsection{Enrollment-PnP Relative Translation}
\label{sec:pnptranslation}

For translation, the enrollment image \(I_e\) serves as the real
anchor. It fixes the subject under the target camera and provides a
metric camera ray through landmark PnP. We detect 68 2D landmarks in
\(I_e\), pair them with the landmarks of the
mean-neutral FLAME head, with zero shape and expression parameters at the
model's metric scale, and solve an
EPnP-RANSAC problem using the RGB intrinsics to obtain the enrollment
translation:

\begin{equation}
    \widehat{\transl}^{\mathrm{PnP}}_e
    =
    [\widehat{x}^{\mathrm{PnP}}_e,
     \widehat{y}^{\mathrm{PnP}}_e,
     \widehat{z}^{\mathrm{PnP}}_e]^\top .
\end{equation}

\noindent The query is then paired with the enrollment anchor:
\begin{equation}
  (\widehat{\Delta\rot}_{q\leftarrow e},
   \widehat{\boldsymbol\delta}_{q\leftarrow e})
  =
  f_\theta(I_e,I_q).
\end{equation}
For the final method, we use only the predicted relative translation update from this real-to-real pair. The query translation is decoded using the enrollment-PnP translation:
\begin{equation}
  \widehat{\transl}^{\,\mathrm{PnP\text{-}rel}}_q
  =
  \operatorname{Decode}
  \left(
  \widehat{\boldsymbol\delta}_{q\leftarrow e},
  \widehat{\transl}^{\mathrm{PnP}}_e
  \right),
  \label{eq:pnp_rel_translation}
\end{equation}
where \(\operatorname{Decode}\) denotes Eqs.~\eqref{eq:zdecode}-\eqref{eq:xydecode}. This branch requires landmark detection only once, at enrollment, and provides the translation component through one real-to-real relative prediction per query. 

\subsubsection{Final Selected Method}

\label{sec:finalmethod}
We refer to the combination of avatar-bank rotation and enrollment-PnP relative translation as \textbf{ReVA-HPE}. The final query pose is

\begin{equation}
\widehat{\pose}_q = \left(\widehat{\rot}^{\mathrm{bank}}_q,\, \widehat{\transl}^{\,\mathrm{PnP\text{-}rel}}_q\right)
\end{equation}
where each component is sourced from its most reliable estimate: the bank for rotation and enrollment PnP for metric translation. Bank translation is used only internally during pose-consensus RANSAC.

%% file: sec_round_2/4_experiments.tex
\section{Experiments}
\label{sec:experiments}

\begin{table*}[t]
\centering
\scriptsize
\resizebox{0.85\textwidth}{!}{%
\begin{tabular}{@{}lccccccccc@{}}
\toprule
Method
& Yaw $\downarrow$
& Pitch $\downarrow$
& Roll $\downarrow$
& MAE$_r$ $\downarrow$
& $T_x$ $\downarrow$
& $T_y$ $\downarrow$
& $T_z$ $\downarrow$
& MAE$_t$ $\downarrow$
& Pose Supervision \\
\midrule

\multicolumn{10}{@{}l}{\textit{Reported numbers}} \\
\midrule

 Direct 6DoF Regress~\cite{kao2023perspnet}
  & 16.49 & 14.03 & 5.81 & 12.11
  & 62.36 & 85.01 & 366.52 & 171.30 & R \\

Refined Pix2Pose~\cite{kao2023perspnet,park2019pix2pose}
  & 5.75 & 5.06 & 11.23 & 7.35
  & 16.82 & 21.30 & 255.36 & 97.83 & R \\

MICA~\cite{zielonka2022mica}
& 5.40 & 7.17 & 3.80 & 5.46
& 9.32 & 13.66 & 60.13 & 27.70 & M \\

JMLR~\cite{guo2022jmlr}
& 6.31 & 6.17 & 3.72 & 5.40
& 8.66 & 7.27 & 32.63 & 16.19 & M \\

PerspNet~\cite{kao2023perspnet}
& \underline{3.10}
& \textbf{3.37}
& 2.38
& 2.95
& \textbf{4.15}
& \textbf{6.43}
& 46.69
& 19.09
& R \\

TRG~\cite{chun2024trg}
& 3.28
& 3.52
& \underline{1.87}
& \underline{2.89}
& 8.41
& 7.38
& \underline{27.13}
& \underline{14.31}
& R \\

TRG$^\ast$~\cite{chun2024trg}
& \textbf{3.04}
& \underline{3.44}
& \textbf{1.78}
& \textbf{2.75}
& \underline{7.83}
& \underline{6.99}
& \textbf{24.07}
& \textbf{12.97}
& M \\

OPAL~\cite{cobo2024opal}
& 4.54 & 5.05 & 2.80 & 4.13
& -- & -- & -- & -- & M \\

\midrule
\multicolumn{10}{@{}l}{%
\textit{Reproduced under shared MTCNN protocol}} \\
\midrule

6DRepNet~\cite{hempel2022sixdrepnet}
& 3.74 & 4.95 & 3.04 & 3.91
& -- & -- & -- & -- & M \\

TokenHPE-v1~\cite{zhang2023tokenhpe}
& 5.57 & 6.23 & 3.79 & 5.20
& -- & -- & -- & -- & M \\

TRG$^\ast$~\cite{chun2024trg}
& 4.58 & 7.18 & 3.68 & 5.15
& 13.03 & 13.59 & 119.52 & 48.71 & M \\

SemiUHPE~\cite{zhou2026semiuhpe}
& 3.30
& 3.71
& \underline{2.60}
& 3.20
& -- & -- & -- & -- & R \\

OPAL~\cite{cobo2024opal}
& 3.74 & 7.23 & 3.25 & 4.74
& -- & -- & -- & -- & M \\

VGGT-HPE-Abs~\cite{vasileiou2026vggt}
& 4.90 & 7.01 & 3.53 & 5.15
& -- & -- & -- & -- & S \\

PnP-enrollment + Rel.
& 2.27
& 4.19
& 3.41
& 3.29
& \textbf{8.43}
& \textbf{11.14}
& \textbf{31.74}
& \textbf{17.10}
& S \\

ReVA-HPE (ours, first frame)
& \underline{2.20}
& \underline{2.94}
& \textbf{2.47}
& \underline{2.53}
& \textbf{8.43}
& \textbf{11.14}
& \textbf{31.74}
& \textbf{17.10}
& S \\

ReVA-HPE (ours, $\alpha \leq 15^\circ$)
& \textbf{2.01}
& \textbf{2.84}
& 2.61
& \textbf{2.49}
& \underline{8.77}
& \underline{11.73}
& \underline{36.34}
& \underline{18.94}
& S \\

\midrule
\multicolumn{10}{@{}l}{%
\textit{Methods using a GT anchor pose}} \\
\midrule

VGGT-HPE$^\dagger$~\cite{vasileiou2026vggt}
& 2.23 & 3.07 & 3.38 & 2.89
& 47.32 & 20.93 & 62.21 & 43.48
& S \\

ReVA-HPE-GT$^\dagger$ (ours)
& \textbf{1.60}
& \textbf{1.92}
& \textbf{2.72}
& \textbf{2.08}
& \textbf{7.53}
& \textbf{6.31}
& \textbf{14.86}
& \textbf{9.57}
& S \\

\bottomrule
\end{tabular}%
}

\makebox[\textwidth][r]{\footnotesize
$^\dagger$ Requires the ground-truth pose of the enrollment image
at test time.}


\caption{
BIWI comparison.
Rotation is in degrees and translation in millimeters, lower is better.
Reported values follow~\cite{chun2024trg}, reproduced results use the
shared MTCNN protocol.
``First frame'' fixes each subject's first available frame as enrollment,
without pose-based selection.
\textbf{Bold} and \underline{underlined} denote the best and second-best
results within each non-oracle block.
The GT-anchor block is ranked independently, with its best values in
\textbf{bold}.
S, R, and M denote synthetic, real, and mixed pose supervision.
}
\vspace{-0.5cm}

\label{tab:biwi_6dof_results}
\end{table*}

\textbf{Datasets and Protocols.}
The BIWI Kinect Head Pose Database~\citep{fanelli2013biwi} contains 24 calibrated RGB-D recordings of 20 individuals with frame-level 6DoF head poses. We treat each recording as a separate subject. For every subject, the first available RGB frame is fixed as enrollment before any score is computed and removed from the query set, leaving 15,654 queries. We additionally evaluate on the \famos \cite{TMPEH:CVPR:2023} test set to assess transfer beyond BIWI. We use only camera index 0 throughout the evaluation. Frame-level 6DoF labels are obtained from the provided ground-truth registrations: for each subject, we rigidly align its identity-specific FLAME-topology template to each registered frame using Procrustes analysis, and use the resulting transformation as the head pose in the dataset camera frame. We follow the same protocol as BIWI, using the first available frame for enrollment and the remaining frames as queries.

\noindent\textbf{Metrics.}
As earlier works, for orientation, we report yaw, pitch, and roll
mean absolute error (MAE) and their average, MAE$_r$. For translation,
we report the per-axis mean absolute errors and their arithmetic mean,
MAE$_t$, in millimeters.

\noindent\textbf{Implementation Details.}
Following the literature~\citep{hempel2022sixdrepnet,zhang2023tokenhpe},
we detect face crops with MTCNN~\citep{zhang2016mtcnn} (We also include an ablation study using RetinaFace\cite{deng2020retinaface} in the Suppl. Mat.) FaceLift~\citep{lyu2024facelift} is used with its released pretrained model trained only on synthetic data to create the personalized avatar. 
We initialize the frozen backbone from the released general-purpose VGGT checkpoint and adapt its camera branch through LoRA. The MTCNN and the FAN-style landmark detector \cite{bulat2017far} use released checkpoints pretrained on real face data.
We use
EPnP~\citep{lepetit2009epnp} inside 300-iteration
RANSAC~\citep{fischler1981ransac}, with a 5-pixel threshold, confidence
0.999, and at least six inliers, followed by iterative inlier refinement. We additionally evaluate the enrollment EPnP-RANSAC reprojection threshold using synthetic landmark projections with controlled noise and outliers. The fixed 5-pixel threshold lies at the tighter end of a stable 5–10 pixel region, the full sensitivity analysis appears in the Suppl. Mat. Neither the pose-consensus nor EPnP sensitivity analysis uses BIWI or FaMoS labels for parameter selection.

\begin{table*}[t]
\centering
\scriptsize
\resizebox{0.85\textwidth}{!}{%
\begin{tabular}{@{}lccccccccc@{}}
\toprule
Method
& Yaw $\downarrow$
& Pitch $\downarrow$
& Roll $\downarrow$
& MAE$_r$ $\downarrow$
& $T_x$ $\downarrow$
& $T_y$ $\downarrow$
& $T_z$ $\downarrow$
& MAE$_t$ $\downarrow$
& Pose Supervision \\
\midrule

\multicolumn{10}{@{}l}{%
\textit{Reproduced under shared MTCNN protocol}} \\
\midrule

6DRepNet~\cite{hempel2022sixdrepnet}
& 4.54 & 8.03 & 5.36 & 5.98
& -- & -- & -- & -- & M \\

TokenHPE-v1~\cite{zhang2023tokenhpe}
& 8.62 & 5.69 & 4.46 & 6.26
& -- & -- & -- & -- & M \\

TRG*~\cite{chun2024trg}
& 6.18 & 6.26 & 4.32 & 5.59
& \textbf{7.52}
& \textbf{5.06}
& 91.40
& 34.66
& M \\

SemiUHPE~\cite{zhou2026semiuhpe}
& 1.98
& \underline{3.65}
& \underline{2.94}
& \underline{2.86}
& -- & -- & -- & -- & R \\

OPAL~\cite{cobo2024opal}
& 8.33 & 5.38 & 5.14 & 6.28
& -- & -- & -- & -- & M \\

VGGT-HPE-Abs~\cite{vasileiou2026vggt}
& 34.41 & 10.60 & 9.73 & 18.29
& 27.14 & 9.15 & 85.10 & 40.46
& S \\

PnP-enrollment + Rel.
& 9.32 & 35.08 & 15.62 & 20.01
& 20.73
& 11.82
& \underline{54.83}
& \underline{29.13}
& S \\

ReVA-HPE (ours, first frame)
& \underline{1.34}
& 6.69
& 3.31
& 3.78
& 20.73
& 11.82
& \underline{54.83}
& \underline{29.13}
& S \\

ReVA-HPE (ours, $\alpha \leq 15^\circ$)
& \textbf{1.21}
& \textbf{3.39}
& \textbf{2.80}
& \textbf{2.47}
& \underline{11.18}
& \underline{7.49}
& \textbf{49.15}
& \textbf{22.61}
& S \\

\midrule
\multicolumn{10}{@{}l}{%
\textit{Methods using a GT anchor pose}} \\
\midrule

VGGT-HPE-Rel$^\dagger$~\cite{vasileiou2026vggt}
& 2.93 & 4.91 & 5.47 & 4.44
& 24.85 & 15.00 & 33.68 & 24.51
& S \\

ReVA-HPE-GT$^\dagger$ (ours)
& \textbf{1.38}
& \textbf{1.77}
& \textbf{2.70}
& \textbf{1.95}
& \textbf{12.43}
& \textbf{3.68}
& \textbf{13.49}
& \textbf{9.87}
& S \\

\bottomrule
\end{tabular}%
}

\makebox[\textwidth][r]{\footnotesize
$^\dagger$ Requires the ground-truth pose of the enrollment image
at test time.}

\par\vspace{-2.5mm}

\caption{
FaMoS comparison under the shared MTCNN protocol.
Rotation is in degrees and translation in millimeters, lower is better.
``First frame'' uses each subject's first available frame as enrollment,
without pose-based selection.
In the non-oracle block, \textbf{bold} and \underline{underlined} denote
the best and second-best results. The GT-anchor block is ranked
independently, with its best values in \textbf{bold}.
S, R, and M denote synthetic, real, and mixed pose supervision.
}
\vspace{-0.5cm}

\label{tab:famos_6dof_results}
\end{table*}

\noindent \textbf{Baselines.}
  We compare against head pose estimation methods spanning both
  rotation-only and 6DoF approaches: 6DRepNet~\cite{hempel2022sixdrepnet},
  OPAL~\cite{cobo2024opal}, SemiUHPE~\cite{zhou2026semiuhpe},
  PerspNet~\cite{kao2023perspnet}, and TRG~\cite{chun2024trg}. 
  All baselines are trained on real-world data or mixed regimes
  (e.g., real images expanded via synthetic pose profiling such as
  300W-LP~\cite{zhu2016facealignment},
  ARKitFace~\cite{kao2023perspnet}, or combinations thereof). 
  We also evaluate \textbf{PnP-enrollment+Rel.}, which uses the same relative network but directly uses the
enrollment-PnP rotation, without virtual-anchor consensus. This serves as the direct ablation
  of the bank's contribution to rotation.

\subsection{Quantitative Results}
\label{sec:results}

Tables~\ref{tab:biwi_6dof_results} and~\ref{tab:famos_6dof_results} present quantitative results on BIWI and FaMoS datasets.

\noindent \textbf{BIWI.}
We report results from the original publications for context, together with results reproduced on the same test frames using a shared preprocessing pipeline. For the reproduced comparison, we use MTCNN because it follows the established BIWI evaluation protocol of 6DRepNet and subsequent work~\cite{hempel2022sixdrepnet,vasileiou2026vggt}. In the Suppl. Mat., we verify that this choice does not favor ReVA-HPE by repeating the evaluation with RetinaFace and showing that the detector choice minimally changes method results.

Our default first-frame protocol fixes each subject's first available frame as
the enrollment image, without pose-based selection. We additionally report a
guided protocol using an enrollment image with
\(\alpha=\max(|\mathrm{yaw}|,|\mathrm{pitch}|,|\mathrm{roll}|)\leq15^\circ\).
Ground-truth angles are used only to construct this diagnostic protocol and are never supplied to ReVA-HPE. In practice, this condition can be encouraged simply by asking the user to face the camera while providing the enrollment image. A detailed enrollment-pose ablation across all angle ranges is presented later.

Under the shared protocol, the default first-frame ReVA-HPE obtains
\(2.53^\circ\), outperforming all reproduced competing baselines, the next best
is SemiUHPE at \(3.20^\circ\). With the guided
\(\alpha\leq15^\circ\) enrollment protocol, ReVA-HPE improves further to
\(2.49^\circ\). Both results are also numerically below the best separately
reported result, TRG$^\ast$ at \(2.75^\circ\), although these cross-protocol
numbers are provided only for context.

The comparison with PnP-enrollment+Rel.\ shows where this improvement comes
from. Both variants use the same relative pose estimator and the same
translation branch. The difference is how query rotation is obtained:
PnP-enrollment+Rel.\ relies on a single PnP estimate of the enrollment
rotation, whereas ReVA-HPE combines evidence from the virtual-anchor bank.
This reduces rotation error from \(3.29^\circ\) to \(2.53^\circ\), showing that
the bank provides a more reliable query-rotation estimate. Their translation
results are identical by design for the same enrollment image because both use
the same enrollment-PnP translation and relative translation decoder.

Translation remains the more difficult part of the problem. The separately
reported TRG$^\ast$ result, which uses mixed pose supervision, has lower
translation error than ReVA-HPE, although this comparison is not
protocol-matched. Under the shared protocol, ReVA-HPE remains competitive
without real pose supervision. Finally, the GT-anchor variant reduces rotation
error further to \(2.08^\circ\). This result measures the remaining cost of
estimating the anchor automatically and indicates that improved enrollment
calibration could further improve performance.

\noindent \textbf{FaMoS.}
FaMoS is particularly challenging because camera~0 is positioned below the
subject. This introduces a strong pitch bias in the PnP estimate obtained from
the enrollment image. As a result, PnP-enrollment+Rel.\ reaches a pitch error of
\(35.08^\circ\) and an overall rotation error of \(20.01^\circ\). ReVA-HPE
avoids using this PnP rotation and instead estimates query rotation by combining
predictions from multiple virtual anchors. Under the default first-frame
protocol, this reduces the rotation error to \(3.78^\circ\), outperforming all
reproduced non-oracle baselines except SemiUHPE, which achieves
\(2.86^\circ\).
Under the guided \(\alpha\leq15^\circ\) protocol defined above, ReVA-HPE
improves further to \(2.47^\circ\). As shown later in
Table~\ref{tab:enrollment_pose_sensitivity}, SemiUHPE obtains
\(3.00^\circ\) on exactly the same evaluation cases. ReVA-HPE therefore
achieves the best rotation accuracy in this guided protocol while using only
synthetic pose supervision for its task-specific adaptation. The same table
provides a detailed analysis across the remaining enrollment-pose ranges.

The benefit is also clear for translation. Under the default first-frame
protocol, ReVA-HPE achieves an MAE$_t$ of \(29.13\)~mm, compared with
\(34.66\)~mm for TRG and \(40.46\)~mm for VGGT-HPE-Abs. Under the guided
protocol, its translation error decreases further to \(22.61\)~mm.

Finally, the ground-truth enrollment-pose oracle reaches \(1.95^\circ\),
showing that the relative predictor generalizes strongly to FaMoS and that much
of the remaining rotation error comes from estimating the enrollment reference
automatically.

Qualitative results of full 6DoF pose on BIWI and FaMoS visualizing both rotation and metric translation for all compared methods are included in the Suppl. Mat.

\subsection{Ablation Studies}
\noindent\textbf{Bank size and efficient inference.}
Table~\ref{tab:bank_size_runtime} compares direct inference and
VA-fast. Direct-1r evaluates $M$ anchors at one radius, Direct-5r
evaluates $5M$ candidates at five radii, and VA-fast uses the bank
once at enrollment followed by one real-to-real prediction per query.
Accuracy largely saturates at $M=20$: increasing Direct-5r to $M=50$
changes MAE$_r$ from $2.53^\circ$ to $2.51^\circ$, while model time
increases from 9.774 to 30.877~s. We therefore use $M=20$.
Avatar reconstruction and bank rendering are one-time enrollment
costs.

\begin{table}[htpb]
\centering
\scriptsize
\setlength{\tabcolsep}{5pt}
\renewcommand{\arraystretch}{0.7}

\begin{tabular}{@{}lccc@{\hspace{4pt}}ccc@{}}
\toprule
& \multicolumn{3}{c}{$M=5$}
& \multicolumn{3}{c}{$M=10$} \\
\cmidrule(lr){2-4}
\cmidrule(lr){5-7}

Protocol
& MAE$_r$
& Time
& Spd.
& MAE$_r$
& Time
& Spd. \\
\midrule

Direct-1r
& 4.30
& 0.683
& --
& 2.92
& 1.165
& -- \\

Direct-5r
& \textbf{3.23}
& 2.472
& --
& \textbf{2.79}
& 4.902
& -- \\

VA-fast
& 4.53
& 0.137
& 5.0/18.0
& 3.08
& 0.117
& 10.0/41.9 \\

\midrule

& \multicolumn{3}{c}{$M=20$}
& \multicolumn{3}{c}{$M=50$} \\
\cmidrule(lr){2-4}
\cmidrule(lr){5-7}

Direct-1r
& 2.69
& 2.185
& --
& 2.58
& 5.259
& -- \\

Direct-5r
& \textbf{2.53}
& 9.774
& --
& \textbf{2.51}
& 30.877
& -- \\

VA-fast
& 2.85
& 0.109
& 20.0/89.7
& 2.97
& 0.105
& 50.1/294.1 \\

\bottomrule
\end{tabular}

\caption{
MAE$_r$ is in degrees and time is model-only seconds per query.
VA-fast speedups are relative to Direct-1r/Direct-5r.}
\vspace{-0.2cm}
\label{tab:bank_size_runtime}
\end{table}




\noindent\textbf{Enrollment sensitivity.}
Tables~\ref{tab:enrollment_pose_sensitivity} and
\ref{tab:enrollment_robustness} study complementary sources of
enrollment error. Across the diagnostic pose bins, ReVA-HPE rotation
degrades as the enrollment becomes more off-axis, gradually on BIWI
and more strongly on FaMoS. SemiUHPE is query-only and is therefore
unchanged by enrollment selection: ReVA-HPE is strongest for
near-frontal enrollment, while SemiUHPE becomes stronger in the
off-axis bins. Table~\ref{tab:enrollment_robustness}(a) applies each
image perturbation before face detection and reruns the complete
pipeline, including FaceLift reconstruction, landmark processing,
PnP, virtual-bank generation, consensus, and query inference.
Downsampling has little effect, blur causes moderate degradation, and
partial occlusion is the most damaging condition, particularly for
metric translation. Panels~(b--c) separately isolate the
enrollment-PnP branch by perturbing only the 68 enrollment landmarks:
mild jitter and random landmark removal are tolerated, whereas larger
coordinate errors primarily corrupt the estimated depth, bank rotation
remains unchanged by construction. The Suppl. Mat. provides
the complete protocols, GT-use policy, all perturbation severities,
coverage and failure accounting, consensus-inlier statistics, and the
full landmark-jitter and removal results.

\begin{table}[thbp]
\centering
\scriptsize
\setlength{\tabcolsep}{2.2pt}
\resizebox{0.9\columnwidth}{!}{%
\begin{tabular}{@{}lcc@{\hspace{7pt}}lcc@{\hspace{7pt}}lcc@{}}
\toprule
\multicolumn{3}{c}{\textbf{(a) Full-pipeline degradation}} &
\multicolumn{3}{c}{\textbf{(b) Landmark jitter}} &
\multicolumn{3}{c}{\textbf{(c) Landmark removal}} \\
\cmidrule(r){1-3}
\cmidrule(lr){4-6}
\cmidrule(l){7-9}

Condition
& MAE$_R$ $\downarrow$
& MAE$_t$ $\downarrow$
&
Condition
& $T_z$ $\downarrow$
& MAE$_t$ $\downarrow$
&
Condition
& $T_z$ $\downarrow$
& MAE$_t$ $\downarrow$ \\
\midrule

Clean
& 2.53
& 17.10
&
Clean
& 31.74
& 17.10
&
Clean
& 31.74
& 17.10
\\

Blur, $\sigma=2$
& 3.22
& 18.64
&
$\sigma=1$ px
& 31.96
& 17.21
&
10\%
& 32.52
& 17.41
\\

Downsample 50\%
& 2.77
& 15.97
&
$\sigma=2$ px
& 31.92
& 17.40
&
25\%
& 32.59
& 17.44
\\

Occlusion 12.5\%
& 4.46
& 38.46
&
$\sigma=4$ px
& 45.55
& 22.35
&
40\%
& 33.83
& 17.97
\\


\bottomrule
\end{tabular}%
}

\caption {Sensitivity to enrollment-image degradation and landmark
errors on BIWI.}
\vspace{-0.6cm}
\label{tab:enrollment_robustness}
\end{table}

\begin{table}[htbp]
\centering
\scriptsize
\setlength{\tabcolsep}{3pt}
\resizebox{0.8\columnwidth}{!}{%
\begin{tabular}{@{}llcccc@{}}
\toprule
Dataset
& Enrollment pose
& Coverage
& \shortstack{ReVA-HPE\\MAE$_r$ $\downarrow$}
& \shortstack{SemiUHPE\\MAE$_r$ $\downarrow$}
& \shortstack{ReVA-HPE\\MAE$_t$ $\downarrow$} \\
\midrule

\multirow{4}{*}{BIWI}
& $\alpha \leq 15^\circ$
& 99.85\%
& \textbf{2.49}
& 3.20
& 18.94 \\

& $15^\circ < \alpha \leq 30^\circ$
& 99.85\%
& 3.71
& \textbf{3.20}
& 18.62 \\

& $30^\circ < \alpha \leq 45^\circ$
& 99.85\%
& 4.85
& \textbf{3.20}
& 19.31 \\

& $\alpha > 45^\circ$
& 99.85\%
& 4.80
& \textbf{3.20}
& 21.77 \\

\midrule

\multirow{4}{*}{FaMoS}
& $\alpha \leq 15^\circ$
& 84.57\%
& \textbf{2.47}
& 3.00
& \textbf{22.61} \\

& $15^\circ < \alpha \leq 30^\circ$
& 73.94\%
& 6.02
& \textbf{2.39}
& 26.83 \\

& $30^\circ < \alpha \leq 45^\circ$
& 55.32\%
& 10.04
& \textbf{2.40}
& 26.59 \\

& $\alpha > 45^\circ$
& 0\%
& --
& --
& -- \\

\bottomrule
\end{tabular}%
}

\caption{
Sensitivity to enrollment pose on BIWI and FaMoS under the shared
MTCNN protocol.}
\vspace{-0.6cm}
\label{tab:enrollment_pose_sensitivity}
\end{table}

\section{Limitations and Discussion}
\label{sec:limitations}

Our results lead to three main observations. First, synthetic pose supervision transfers well in the relative formulation: ReVA-HPE shows substantially less cross-dataset degradation than the matched synthetic absolute baseline. Domain shift nevertheless remains, as performance still depends on the dataset, anchor geometry, avatar quality, and camera calibration. Second, the virtual-anchor bank remains effective even though no rendered view perfectly matches the query. Consensus across virtual anchors suppresses inconsistent hypotheses and improves rotation accuracy. Third, increasing the bank size helps up to \(M{=}20\), after which performance plateaus or declines as additional virtual anchors extend farther from the frontal pose range.

Metric translation still depends on PnP applied to the real anchor image. This provides a simple and effective metric reference, but makes translation sensitive to landmark quality and enrollment pose. Extreme orientations can degrade landmark localization and the resulting PnP estimate. The real anchor should therefore preferably have a near-neutral pose, although moderate off-axis poses are well tolerated.

The main remaining limitation is the dependence on avatar and calibration quality. Errors in identity, hair, silhouette, or unseen profile geometry can weaken anchor consensus. Moreover, avatar reconstruction, landmark-based calibration, and relative pose estimation are separate stages and are not jointly optimized. Future work could optimize the avatar directly for geometric matching and exploit temporal continuity in video, since the current method treats each query independently.

%% file: sec_round_2/7_conclusion.tex
\section{Conclusion}
\label{sec:conclusion}


We presented ReVA-HPE, which extends relative head pose estimation to complete metric 6DoF without requiring a ground-truth anchor pose at inference. The pose estimator uses only 25,000 procedural FLAME renderings, without real-image pose supervision, pseudo-labels, or target-domain fine-tuning. At inference, a personalized virtual-anchor bank provides robust rotation hypotheses through multi-anchor consensus, while a single PnP solve on the enrollment image grounds metric translation. On BIWI, ReVA-HPE achieves \(2.53^\circ\) rotation MAE and outperforms all baselines reproduced under the same protocol, including methods using real or mixed pose supervision.

\section*{Acknowledgments}
\begingroup
\small
\sbox{\acklogo}{%
  \includegraphics[height=2em]{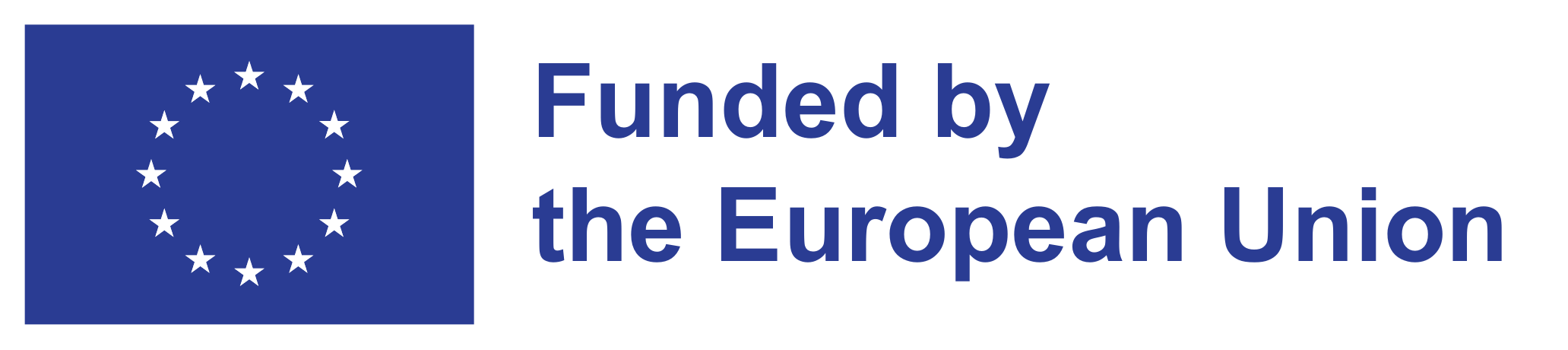}%
}
\noindent
\hangindent=\dimexpr\wd\acklogo+8pt\relax
\hangafter=-2
\llap{%
  \raisebox{\dimexpr\ht\strutbox-\ht\acklogo\relax}[0pt][0pt]{%
    \usebox{\acklogo}%
  }%
  \hspace{8pt}%
}%
This project is funded by the European Union under Horizon Europe (grant No.\ 101136568 -- HERON). V. Vasileiou was also supported by a PhD fellowship from Archimedes/Athena RC. We acknowledge EuroHPC Joint Undertaking for awarding us access to Leonardo at CINECA, Italy, under project EHPC-DEV-2026D01-089.
\par
\endgroup
